\documentclass[conference]{IEEEtran}
\IEEEoverridecommandlockouts
\usepackage{cite}
\usepackage{amsmath,amssymb,amsfonts}
\usepackage{algorithmic}
\usepackage{graphicx}
\usepackage{textcomp}
\usepackage{xcolor}
\usepackage{booktabs}
\usepackage{tabularx}
\usepackage{array}
\usepackage{url}
\usepackage{hyperref}

\def\BibTeX{{\rm B\kern-.05em{\sc i\kern-.025em b}\kern-.08em
    T\kern-.1667em\lower.7ex\hbox{E}\kern-.125emX}}
\begin{document}

\title{Thermal Imaging for Contactless Cardiorespiratory and Sudomotor Response Monitoring
}

\author{
    \IEEEauthorblockN{
        Constantino \'Alvarez Casado\IEEEauthorrefmark{1}\IEEEauthorrefmark{2},
        Mohammad Rahman\IEEEauthorrefmark{1},
        Sasan Sharifipour\IEEEauthorrefmark{1},
        \\
        Nhi Nguyen\IEEEauthorrefmark{1},
        Manuel Lage Ca\~nellas\IEEEauthorrefmark{1},
        Xiaoting Wu\IEEEauthorrefmark{1},
        Miguel Bordallo L\'opez\IEEEauthorrefmark{1}
    }
    \vspace{2mm}
    \IEEEauthorblockA{\IEEEauthorrefmark{1}\textit{Center for Machine Vision and Signal Analysis (CMVS), University of Oulu, Finland}}

    \IEEEauthorblockA{\IEEEauthorrefmark{2}\textit{Candour Ltd, Oulu, Finland}}
}

\maketitle

\begin{abstract}

Human-machine interfaces in industrial automation need sensing modules that can monitor not only what operators do, but also their physiological state. This is important in factories, vehicles, machinery cabins, and human-robot collaboration, where workload, stress, fatigue, or reduced attention can affect safety and decision-making. RGB-based monitoring is limited by low light, shadows, and privacy concerns, while thermal infrared imaging captures skin temperature dynamics without visible illumination. This paper studies thermal video as a contactless computer vision modality for estimating electrodermal activity (EDA), heart rate (HR), and breathing rate (BR), with the long-term goal of supporting adaptive human-machine interfaces and operator-state awareness. We propose a signal-processing pipeline that tracks facial anatomical regions, aggregates thermal signals, and separates slow \textcolor{black}{sudomotor (sweat-gland-related)} trends from faster cardiorespiratory components. HR is estimated using orthogonal matrix image transformation (OMIT) across multiple facial regions, while BR is estimated from nasal and cheek thermal signals using spectral peak detection. \textcolor{black}{The study reports a unified characterization of 288 ROI--method configurations against synchronized contact references with lag-tolerant metrics.} The evaluation uses 31 sessions from the public SIMULATOR STUDY 1 (SIM1) driver monitoring dataset. The best fixed EDA configuration reaches a mean absolute correlation of $0.40 \pm 0.23$ against palm EDA, with individual sessions reaching $0.89$. BR estimation achieves $3.1 \pm 1.1$\,bpm mean absolute error, while HR estimation yields $13.8 \pm 7.5$\,bpm MAE, partly limited by the $7.5$\,Hz thermal camera frame rate. The results show that thermal video can provide useful respiratory and sudomotor cues, while also revealing limitations caused by region selection, polarity changes, latency, and subject variability. These findings provide \textcolor{black}{a baseline design guidance for thermal computer vision as an auxiliary sensing layer in adaptive industrial HMI systems.} The code is available at: \href{https://github.com/Multimodal-Sensing-Lab/oulu-human-thermal-sensing}{https://github.com/Multimodal-Sensing-Lab/oulu-human-thermal-sensing}.

\end{abstract}


\section{Introduction}

\begin{figure}[ht!]
\centering
\includegraphics[width=\linewidth]{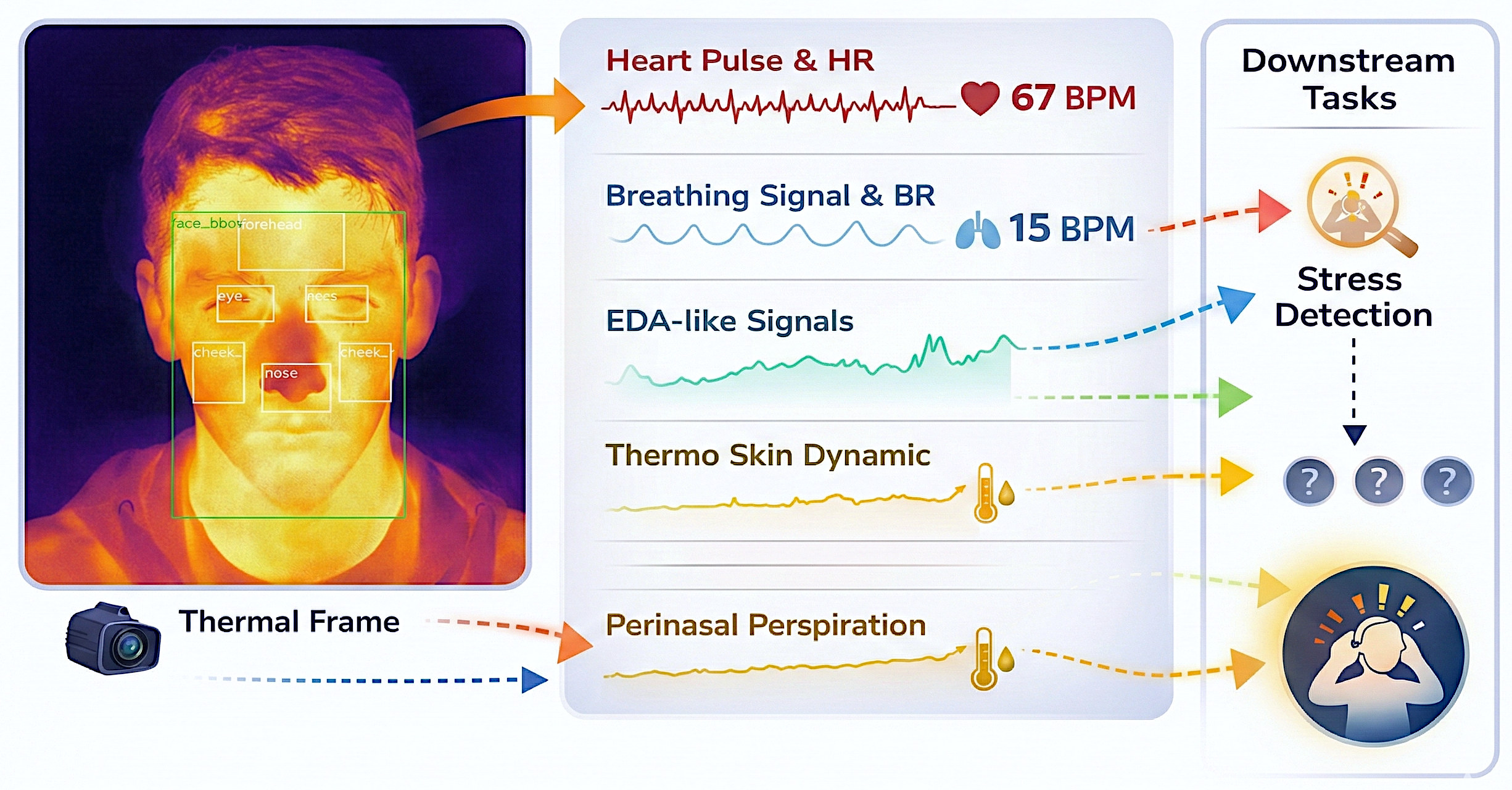}
\vspace{-6mm}
\caption{Thermal facial video can be converted into ROI temperature traces and decomposed into respiratory, cardiac-related, and slow autonomic components, including EDA-like trends and perinasal perspiration proxies.}
\label{fig:thermal_biosignals_overview}
\end{figure}

\textcolor{black}{Industrial and factory automation systems are progressively moving from machine-centered control toward human-centered interaction~\cite{xu2026human}. Operators increasingly supervise autonomous systems, interact with collaborative robots, operate vehicles or machinery, and make decisions under time pressure. In this context, human-machine interfaces (HMIs) are expected to adapt not only to external actions, but also to the cognitive and physiological state of the operator~\cite{meftah2026adaptiveHMI,antonaci2024workplace}. In high-stakes tasks, measurements such as breathing rate, heart rate, and EDA-related trends can provide interpretable evidence of operator state, complementing black-box labels such as stress, fatigue, or cognitive overload~\cite{rudin2019stop}, as shown in Figure \ref{fig:thermal_biosignals_overview}.  Contactless monitoring is therefore relevant in safety-critical settings including industrial work, driving, and human-robot collaboration~\cite{LI2026132877,wang2025contactlessMachinery}, where skin-contact sensors are impractical during long shifts or active tasks~\cite{Rubin2024AItherapy}.}


\vspace{-3mm}

\textcolor{black}{Most contactless approaches estimate heart rate (HR) and breathing rate (BR) from RGB video or radio-frequency sensing~\cite{LI2026132877,alvarez2022face2ppg,nguyen2022identification}. However, RGB-based monitoring can be degraded by low light, shadows, reflective surfaces, dust, smoke, and partial occlusions, and facial video may raise privacy concerns when the objective is physiological monitoring rather than identity recognition. Estimating electrodermal activity (EDA) from RGB or radio-frequency signals is also less direct, since EDA reflects changes in skin conductance driven by eccrine (sweat-gland) activity controlled by the sympathetic nervous system and is conventionally measured with electrodes~\cite{jukiewicz2021electrodermal}. Since EDA is a standard marker of sympathetic activation, its contactless estimation would provide a complementary cue for workload, stress, and arousal analysis in adaptive HMI~\cite{pavlidis2016dissecting}.}


\textcolor{black}{Thermal infrared imaging is a relevant modality for this purpose because it captures cutaneous temperature fields modulated by perfusion, vasomotor (blood-vessel-calibre) tone, sweating, and convective heat exchange~\cite{gade2014thermal,LI2026132877}. Thermal cameras do not depend on visible illumination, have been explored in low-light industrial perception and human-robot collaboration~\cite{sume2025thermalHRC}, and can support embedded respiration monitoring on edge hardware~\cite{analia2026adaptiveThermalRR}. Physiologically, temperature variations around the nostrils and mouth can reflect respiratory airflow, while facial and peripheral thermal changes may contain cardiac-related components and slower sudomotor (sweat-gland-related) dynamics \cite{garbey2007contactfree}. These properties are relevant for driver and operator-state monitoring, where vigilance, workload, and physiological state awareness are important~\cite{cardone2020driverStressThermal,pavlidis2016dissecting}.} 


\textcolor{black}{Despite this potential, thermal physiological monitoring remains affected by sensor noise, ambient drift, airflow, reflections, limited frame rate, thermodynamic latency, head motion, ROI selection, occlusions, and inter-subject variability~\cite{gade2014thermal}. Although thermal facial features have been used for stress, workload, fatigue, and arousal analysis, the extent to which HR, BR, and EDA-related dynamics can be recovered from facial thermal video remains insufficiently characterized when waveform-level agreement, ROI selection, processing choices, task conditions, and participant variability are evaluated together~\cite{xiao2024reading}.}


\textcolor{black}{Motivated by this gap, this paper characterizes interpretable extraction of EDA-like trends and cardiorespiratory components from facial thermal video for operator-state awareness and adaptive HMI. We present a pipeline that detects facial landmarks in thermal frames, defines six anatomical regions of interest (ROIs), extracts ROI temperature traces by spatial aggregation, and applies temporal decomposition to separate slow sudomotor trends, cardiac-related components, and respiratory oscillations. We evaluate the approach on the SIMULATOR STUDY 1 (SIM1) dataset~\cite{taamneh2017SMI1}, covering 31 sessions from eight subjects under four driving conditions. The analysis compares 288 ROI--method configurations for EDA-like trends and a multi-ROI decomposition strategy for HR and BR, reporting agreement with synchronized contact references and examining task, sex, and age effects.} \textcolor{black}{The individual detection, filtering, and spectral-estimation blocks are based on established components; the contribution is the controlled, joint characterization of their use for EDA-like, HR, and BR extraction under one lag-aware protocol.} \textcolor{black}{The contributions of this work are:}

\begin{itemize}
    \item \textcolor{black}{A reproducible thermal benchmark protocol that integrates thermal face landmarking, anatomical ROI aggregation, slow-trend extraction, OMIT-based HR estimation, and nasal/cheek BR estimation on the same SIM1 facial thermal stream.}
    \item \textcolor{black}{A controlled comparison of 288 ROI--method configurations across six anatomical ROIs and eight slow-trend extraction methods, quantifying fixed and per-session ROI behavior, filtering choices, task-dependent variability, and landmark-smoothing sensitivity.}
    \item \textcolor{black}{A protocol-aware comparison with prior thermal BR, HR, and EDA studies, together with analysis of polarity inversion, thermodynamic lag, and performance bounds for SIM1, providing design guidance for adaptive HMI and future learning-based benchmarks.}
\end{itemize}

\vspace{2mm}
\section{Related Work}
\label{sec:relatedwork}

Thermal cameras capture emitted infrared radiation, typically in the LWIR band (7.5--14\,$\mu$m), and convert radiance to apparent temperature through calibration and emissivity assumptions~\cite{LI2026132877,gade2014thermal}. The human skin emissivity is close to that of a blackbody ($\varepsilon \approx 0.98$), so radiometric measurements approximate the temperature of the superficial skin under controlled conditions. 

Respiration is the most established target for thermal biosignal extraction, as inhalation and exhalation create temperature oscillations in the perinasal region. The virtual thermistor approach~\cite{fei2010thermistor} and later extensions have validated breathing-rate estimation against contact sensors in laboratory, neonatal, and clinical settings~\cite{cho2017robust,maurya2023neonatal,aldred2022thermography}, although performance depends on nostril visibility, motion, and thermal contrast~\cite{lorato2020multicamera}. Heart rate estimation is less consistent because pulsatile thermal modulation is weak and filtered by heat diffusion. Vessel-focused methods analyse superficial arteries~\cite{garbey2007contactfree,chekmenev2007thermal}, while motion-proxy methods estimate cardiac micro-motion from tracked facial features~\cite{pereira2018monitoring}; both generally require stable head pose and higher frame rates than standard microbolometers provide. Sudomotor activity is a distinctive opportunity for thermal sensing, since sweat-gland activation produces evaporative cooling detectable in high-resolution thermography. Krzywicki et al.~\cite{krzywicki2014noncontact} detected sweat-pore activation correlated with skin conductance responses, Sagaidachnyi et al.~\cite{sagaidachnyi2022separate} separated sweat-related and \textcolor{black}{haemodynamic (blood-flow-related)} components, and \textcolor{black}{Gioia et al.~\cite{gioia2025thermica} reported high agreement between ICA-derived facial thermal components and contact EDA under constrained laboratory conditions.} These studies support the feasibility of thermal EDA estimation, but further characterization is needed under less constrained settings with multiple facial regions, head motion, task changes, and polarity variations. Thermal facial patterns have also been used to classify stress, cognitive workload, fatigue, and arousal~\cite{xiao2024reading,sonkusare2019detecting}, but these works typically treat temperature as a classification feature rather than reconstructing physiological waveforms.

\textcolor{black}{In industrial HMI, physiological monitoring is motivated by adaptive interfaces, operator-state awareness, and worker well-being in human-centered automation~\cite{meftah2026adaptiveHMI,antonaci2024workplace,xu2026human}. Existing studies cover adjacent parts of the problem rather than the same evaluation target: contactless vital-sign monitoring has been studied for construction machinery operators~\cite{wang2025contactlessMachinery}, thermal vision has been applied to human-robot collaboration in manufacturing~\cite{sume2025thermalHRC}, and embedded thermal systems target local respiratory monitoring~\cite{analia2026adaptiveThermalRR}. In the thermal-physiology literature, BR studies usually focus on perinasal oscillations, HR studies commonly rely on vessel-focused ROIs or controlled high-frame-rate acquisitions, and thermal EDA studies often use high-resolution or constrained recordings~\cite{fei2010thermistor,cho2017robust,garbey2007contactfree,pereira2018monitoring,krzywicki2014noncontact,gioia2025thermica}. However, each of these studies targets a single signal and reports accuracy under acquisition conditions its setup was designed to provide: a high frame rate, high spatial resolution, or a fixed pose that keeps the ROI aligned. None of these holds at 7.5\,Hz with free head motion and changing tasks, so how such methods behave, and how the three signals compare on one thermal stream, has not been quantified. Compared with these works, the present study provides a common protocol that jointly evaluates EDA-like thermal trends, HR, and BR from the same facial thermal stream across six anatomical ROIs, eight slow-trend extraction methods, four tasks, participant groups, polarity inversion, and lag-tolerant metrics.}

\section{Methodology}\label{sec:methodology}

\begin{figure*}[ht!]
    \centering
    \includegraphics[width=\textwidth]{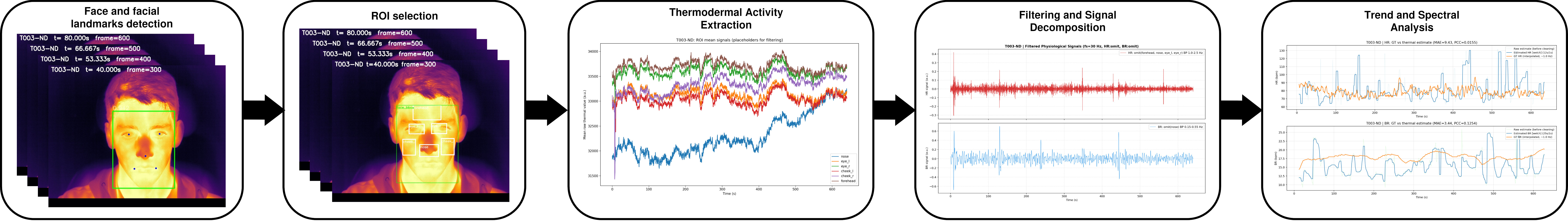}
    \vspace{-6mm}
    \caption{Pipeline overview. Thermal frames are processed by a face detector to localise landmarks. Six facial ROIs are defined and scalar temperature traces are extracted per frame. Temporal decomposition separates slow \textcolor{black}{sudomotor (sweat-gland-related)} trends, cardiac pulse, and respiratory components for comparison with synchronised contact ground truth.}
    \label{fig:pipeline_overview}
    \vspace{-3mm}
\end{figure*}

We propose an interpretable four-stage pipeline for extracting contactless biosignals from thermal infrared video: anatomical ROI detection, spatial aggregation, temporal decomposition into physiological bands, and validation against contact ground truth. Figure~\ref{fig:pipeline_overview} shows the workflow.


\subsection{Detection and ROI Definition}

We use YOLOv5-Face~\cite{qi2022yolo5face} with Thermal Faces in the Wild (TFW) weights~\cite{kuzdeuov2022tfw}, trained on thermal face data from SpeakingFaces~\cite{abdrakhmanova2021speakingfaces}. Raw 16-bit thermal frames ($640 \times 512$ pixels) are converted to 8-bit pseudo-color using a second--98th percentile stretch. The detector returns a face bounding box and five landmarks: left eye, right eye, nose tip, and left/right mouth corners. If several faces are detected, the highest-confidence one is selected. To reduce thermal-noise jitter, each landmark coordinate $p_t$ is smoothed with an exponential moving average, $\hat{p}_t = \alpha p_t + (1-\alpha)\hat{p}_{t-1}$. \textcolor{black}{Unless otherwise stated, we use $\alpha = 0.15$ as a compromise between suppressing frame-to-frame landmark jitter at the 7.5\,Hz native frame rate and preserving responsiveness to natural head motion.} 

From the stabilized landmarks, we define six axis-aligned ROIs proportional to the face bounding box ($b_w \times b_h$), as shown in Figure~\ref{fig:thermal_biosignals_overview}: nose ($0.30\,b_w \times 0.15\,b_h$), centered at the nose tip; left/right periorbital regions ($0.24\,b_w \times 0.12\,b_h$), near the inner canthus, \textcolor{black}{where superficial vasculature and sympathetic thermal responses have been studied}; left/right cheeks ($0.20\,b_w \times 0.20\,b_h$), derived from eye-mouth midpoints with lateral offset; and forehead ($0.45\,b_w \times 0.18\,b_h$), projected above the inter-eye midpoint. All rectangles are clipped to the frame boundaries.

\subsection{Spatial Aggregation}

For each ROI patch $\mathbf{P} \in \mathbb{R}^{h \times w}$, spatial aggregation produces one scalar thermal value per frame. Besides the arithmetic mean, we test a Gaussian-weighted mean (2D kernel with $\sigma = 0.35 \times$ half-width), a trimmed mean discarding the top and bottom 10\% of pixels, and the mean of the hottest 30\% of pixels. \textcolor{black}{Gaussian weighting is used for the nose, cheeks, and forehead to emphasize central skin pixels, while the smaller periorbital ROIs use a trimmed mean to reduce hot/cold outlier effects.}



\subsection{Temporal Decomposition}
\textcolor{black}{ROI signals sampled at the native camera rate (7.5\,Hz for SIM1) are interpolated to a 30\,Hz common time grid by cubic spline interpolation. This resampling does not increase the temporal information content of the thermal video. It is used only to align the thermal ROI traces with the synchronized contact signals and to apply the same temporal filtering pipeline across modalities.} The temporal thermal-based signal is then decomposed into three physiological bands:

\vspace{1mm}
\subsubsection{\textcolor{black}{\textbf{Sudomotor (sweat-gland-related) trends (EDA-like, $<$0.1\,Hz)}}} 
We compare eight extraction methods: Butterworth and Bessel low-pass filters ($f_c = 0.05$\,Hz, 3rd order, zero-phase), Savitzky--Golay filter (30\,s window, 3rd order polynomial), simple and exponential moving averages (30\,s window), median filter followed by Savitzky--Golay smoothing, Hilbert envelope extraction (bandpass 0.05--3\,Hz, amplitude demodulation, then low-pass), and wavelet approximation (Daubechies-4, retaining the low-frequency approximation coefficients). \textcolor{black}{The extracted trends are downsampled to 1\,Hz for metric calculation, since EDA-like dynamics are slow and the higher synchronization rate is unnecessary for correlation analysis.}

\subsubsection{\textbf{Cardiac pulse (1.0--3.5\,Hz, 60--210\,bpm)}} 
Because pulsatile thermal modulation is weak and spatially distributed, we combine multiple ROIs before spectral analysis. The forehead, nose, and bilateral cheek ROI signals are first pre-filtered with a 4th-order Butterworth bandpass (0.3--4.0\,Hz), then combined using an orthogonal matrix image transformation (OMIT)~\cite{alvarez2022face2ppg} decomposition. OMIT performs a QR factorization of the multi-channel signal matrix, projects out the dominant (motion-correlated) component, and preserves the residual channel with the highest spectral peak in the cardiac band. The resulting pulse signal is bandpass-filtered (1.0--3.5\,Hz, 4th order Butterworth) and passed to a sliding-window Welch spectral estimator (15\,s window, 1\,s step) with parabolic peak interpolation. Estimated rates outside the valid range (60--180\,bpm) are replaced by interpolation over short gaps ($\leq$10 samples), followed by a 7-point median filter.

\subsubsection{\textbf{Respiratory oscillation (0.12--0.55\,Hz, 7--33\,bpm)}} The nose and bilateral cheek signals are pre-filtered (0.12--2.0\,Hz bandpass) and averaged to form a combined respiratory signal. The same Welch-based sliding-window estimator is applied (25\,s window, 1\,s step), with valid-range filtering at 7--45\,rpm.

\subsection{Dataset: SIMULATOR STUDY 1 (SIM1)}

\textcolor{black}{The SIMULATOR STUDY 1 (SIM1) dataset~\cite{taamneh2017SMI1} is a public multimodal driving-simulator corpus with a working set of 68 volunteers, after excluding one participant with motion sickness and nine with incomplete recordings. The working set includes 35 male and 33 female subjects from two age cohorts: young drivers between 18 and 27 years, and older drivers above 60 years.} 
\textcolor{black}{Facial thermal video was acquired with a Tau 640 long-wave infrared (LWIR) camera (FLIR Commercial Systems, Goleta, CA) with a 35\,mm f/1.2 LWIR lens, positioned approximately 1.2\,m from the participant. Frames are stored as 16-bit images at $640 \times 512$ pixels and approximately 7.5\,fps.} \textcolor{black}{Although SIM1 also includes visual video, eye tracking, and simulator variables, this work uses only facial thermal video and physiological references.} \textcolor{black}{The synchronized references include palm EDA (PEDA, in k$\Omega$) from a Shimmer3 GSR sensor, HR and BR from a Zephyr BioHarness 3.0 chest strap, and thermal perinasal perspiration signals (PP and noise-reduced PP\_NR, in $^\circ$C$^2$).} \textcolor{black}{The dataset documentation reports that PP signals could not be reliably extracted for nine male subjects because of facial hair, which is relevant for interpreting perinasal ROI quality.}

\textcolor{black}{For demographic subgroup analyses (sex $\times$ age), we use a balanced subset of eight subjects (T002, T003, T005, T014, T029, T031, T034, T036): four young subjects (two female, two male) and four older subjects (two female, two male). This design avoids sex/age imbalance in subgroup comparisons.} Four driving conditions are analyzed per subject: practice drive (PD, $\sim$3\,min), normal driving (ND), cognitive distraction (CD), and emotional distraction (ED), each $\sim$10.7\,min. \textcolor{black}{This yields 31 analyzed sessions, with one session excluded due to a missing synchronization file. Since the pipeline is unsupervised, no training, validation, or test split is required.}


\subsection{Evaluation Metrics}

\textcolor{black}{For EDA-like trends, we report the signed Pearson correlation $r(e,r)$ and its magnitude, $\text{PCC}_{\text{abs}} = |r(e,r)|$, between the estimated thermal trend $e$ and the contact reference $r$. The signed value preserves the direction of the association, while $\text{PCC}_{\text{abs}}$ measures waveform agreement independently of polarity. This is useful because facial thermal trends may reflect different heat-exchange mechanisms, including vasomotor changes and sweat-related cooling~\cite{sagaidachnyi2022separate,gade2014thermal}.} We also compute the Spearman rank correlation $\rho_s$ and the maximum normalized cross-correlation $R_{\max}$ within a conservative $\pm 120$\,s lag window:
\vspace{-2mm}
\begin{equation}
    R_{er}(\tau) = \frac{\sum_i \tilde{e}_{i+\tau}\,\tilde{r}_i}{\sqrt{\sum_i \tilde{e}_i^2}\,\sqrt{\sum_i \tilde{r}_i^2}},
\end{equation}
where $\tilde{e}$ and $\tilde{r}$ are zero-mean unit-variance signals. The lag $\tau^*$ at maximum $R_{er}$ is used to estimate the temporal offset between the thermal signal and the contact reference. \textcolor{black}{The $\pm 120$\,s window is set conservatively to accommodate possible thermodynamic latency from tissue heat transfer and synchronization tolerances inherent to contactless sensing. The lag window is used only for offline characterization of thermal delay and synchronization tolerance; real-time use would require a causal delay-compensation strategy.} We also compute trend agreement, defined as the percentage of time points where the first differences of $e$ and $r$ have the same sign. For HR and BR, we report mean absolute error (MAE), root mean squared error (RMSE), Pearson correlation (PCC), and signed bias, computed on the windowed rate estimates aligned with the contact ground truth.

%
%
%
%

\section{Experimental Results}\label{sec:results}

This section presents the results of applying the proposed pipeline to the SIM1 dataset. We evaluate 288 extraction configurations (6~ROIs $\times$ 8~trend-extraction methods) across 31 subject$\times$task sessions, excluding T005-ND due to a missing synchronization file. Agreement is assessed against two references: the noise-reduced perinasal perspiration signal (PP\_NR), a thermally derived EDA proxy provided with SIM1, and contact palm electrodermal activity (PEDA) measured with a galvanic skin response sensor.



\subsection{EDA: Global Agreement}

\textcolor{black}{Table~\ref{tab:global_results} reports the best fixed ROI--method combinations. Against PEDA, the nose with exponential MA achieves $\text{PCC}_{\text{abs}} = 0.40 \pm 0.23$, followed closely by the cheeks ($0.39 \pm 0.24$). Against PP\_NR, the cheeks lead ($0.32 \pm 0.18$), with the nose and forehead at a similar level.}

\vspace{-4mm}
\begin{table}[ht!]
\centering
\def\arraystretch{1.0}
\setlength{\tabcolsep}{2.4em}
\caption{Top EDA ROI-method combinations ($\text{PCC}_{\text{abs}}$, mean $\pm$ std, $n=31$).}
\vspace{-3mm}
\label{tab:global_results}
\begin{tabular}{@{}llcc@{}}
\toprule
\textbf{ROI} & \textbf{Method} & \textbf{vs PP\_NR} & \textbf{vs PEDA} \\
\midrule
nose       & exp.\ MA     & 0.31$\pm$0.18 & \textbf{0.40$\pm$0.23} \\
cheeks     & exp.\ MA     & \textbf{0.32$\pm$0.18} & 0.39$\pm$0.24 \\
nose       & mov.\ avg.   & 0.30$\pm$0.20 & 0.38$\pm$0.21 \\
forehead   & exp.\ MA     & 0.31$\pm$0.19 & 0.36$\pm$0.25 \\
eye\_r     & exp.\ MA     & 0.28$\pm$0.15 & 0.35$\pm$0.24 \\
cheeks     & mov.\ avg.   & 0.30$\pm$0.18 & 0.34$\pm$0.23 \\
nose       & Bessel LP    & 0.29$\pm$0.18 & 0.34$\pm$0.19 \\
forehead   & mov.\ avg.   & 0.27$\pm$0.19 & 0.32$\pm$0.24 \\
\bottomrule
\end{tabular}
\vspace{-1mm}
\end{table}

\begin{figure}[ht!]
\vspace{-1mm}
    \centering
    \includegraphics[width=\linewidth]{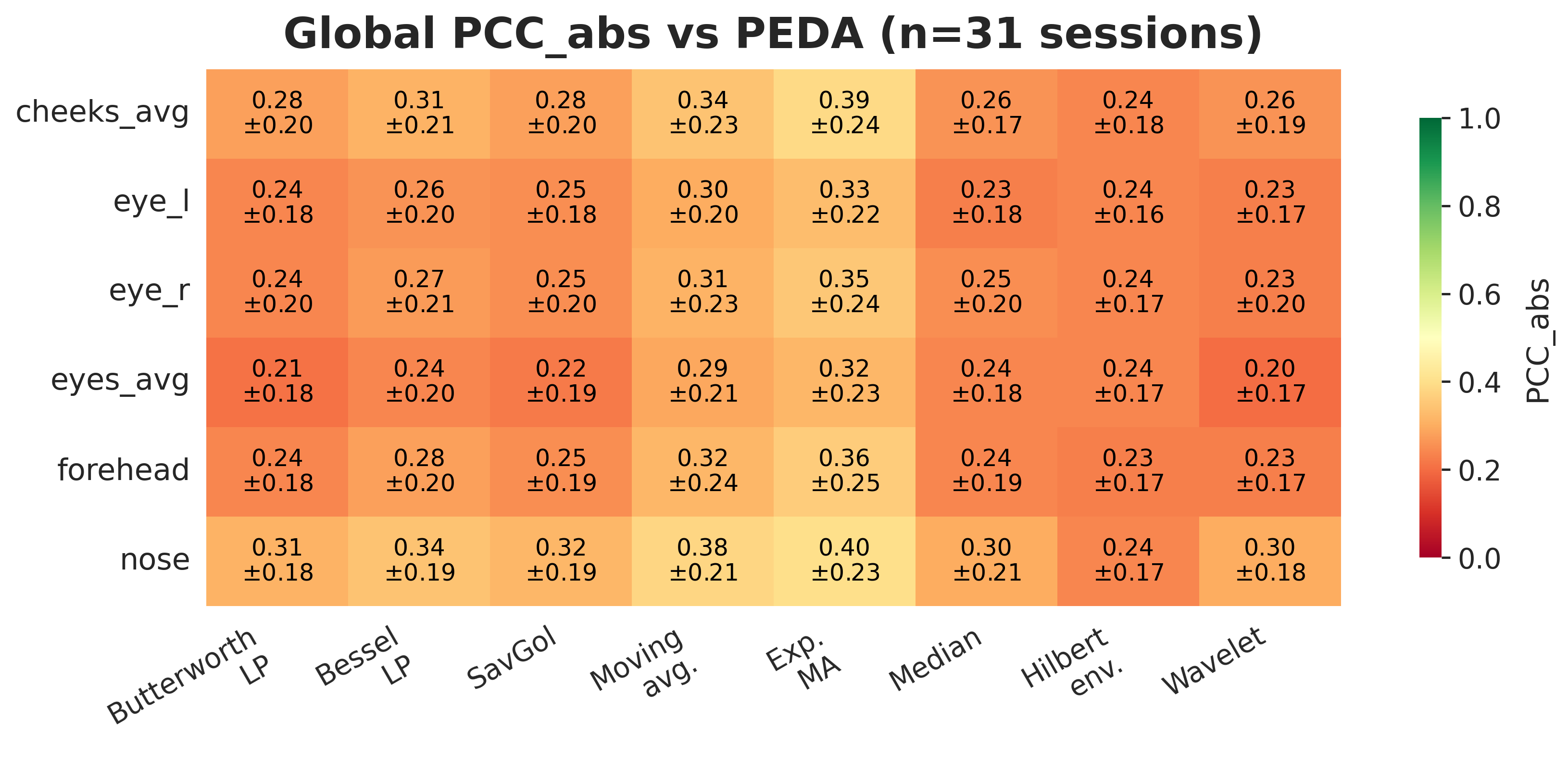}
    \vspace{-8mm}
    \caption{$\text{PCC}_{\text{abs}}$ against PEDA for all 48 ROI-method combinations ($n = 31$).}
    \label{fig:heatmap_global}
    \vspace{-1mm}
\end{figure}

Figure~\ref{fig:heatmap_global} shows the full heatmap against PEDA. Time-domain smoothers, especially exponential and moving averages, outperform the frequency-domain alternatives. In the global aggregation, the strongest regions are the nose and cheeks, while the forehead and periorbital regions become competitive in specific tasks. The small difference between the nose and cheeks, together with the high session-to-session variability (std $\approx 0.23$--$0.24$), indicates that the global mean pools several task- and session-dependent regimes. \textcolor{black}{If we allow per-session oracle selection among the 48 ROI--method combinations, the mean increases to 0.49 against PP\_NR and 0.54 against PEDA, with peaks of 0.78 and 0.89. This oracle result is not intended as a deployable estimator, but it shows that a fixed ROI can miss strong thermal--contact alignment in individual sessions. The fixed nose/exponential-MA mean of $0.40$ is therefore a fixed-configuration baseline; adaptive ROI selection or multi-ROI fusion may recover stronger agreement when local artifacts or polarity affect a single region.} 

\textcolor{black}{Task-stratified results explain part of this variability. Table~\ref{tab:task_top2_eda} reports the two strongest ROI--method combinations per task, computed over the alpha-sensitivity sweep. Against PEDA, PD favours the right periorbital and forehead signals ($\text{PCC}_{\text{abs}}\approx0.58$), ND favours the nose ($\approx0.49$), CD is weaker even for the best nose configuration ($\approx0.26$), and ED favours the cheeks ($\approx0.41$). Against PP\_NR, cheeks lead in PD and ND, forehead in CD, and nose in ED. Therefore, variability is driven not only by ROI choice, but also by driving condition, with CD showing the weakest PEDA agreement and the largest alpha sensitivity.}


\vspace{-5mm}
\begin{table}[ht!]
\def\arraystretch{1.0}
\setlength{\tabcolsep}{0.4em}
\centering
\caption{\textcolor{black}{Top two EDA ROI--method combinations per task. Mean and standard deviation are computed over subject--alpha observations; $\Delta_\alpha$ is the range of the alpha-wise mean $\text{PCC}_{\text{abs}}$.}}
\vspace{-2mm}
\label{tab:task_top2_eda}
\scriptsize
\resizebox{\linewidth}{!}{%
{\color{black}
\begin{tabular}{@{}ccllrrrr@{}}
\toprule
\textbf{Reference} & \textbf{Task} & \textbf{ROI} & \textbf{Method} & \textbf{Rank} & \textbf{Mean} & \textbf{Std} & $\boldsymbol{\Delta_\alpha}$ \\
\midrule
PEDA & PD & eye\_r & exp.\ MA & 1 & 0.5817 & 0.1962 & 0.0180 \\
PEDA & PD & forehead & exp.\ MA & 2 & 0.5782 & 0.1980 & 0.0075 \\
PEDA & ND & nose & median & 1 & 0.4887 & 0.2844 & 0.0170 \\
PEDA & ND & nose & exp.\ MA & 2 & 0.4848 & 0.2705 & 0.0144 \\
PEDA & CD & nose & exp.\ MA & 1 & 0.2610 & 0.1249 & 0.0336 \\
PEDA & CD & nose & mov.\ avg. & 2 & 0.2449 & 0.1111 & 0.0300 \\
PEDA & ED & cheeks & exp.\ MA & 1 & 0.4059 & 0.1756 & 0.0044 \\
PEDA & ED & cheeks & mov.\ avg. & 2 & 0.3580 & 0.1623 & 0.0065 \\
PP\_NR & PD & cheeks & exp.\ MA & 1 & 0.3320 & 0.2189 & 0.0071 \\
PP\_NR & PD & eye\_r & exp.\ MA & 2 & 0.3194 & 0.1776 & 0.0062 \\
PP\_NR & ND & cheeks & exp.\ MA & 1 & 0.4167 & 0.0879 & 0.0374 \\
PP\_NR & ND & forehead & exp.\ MA & 2 & 0.3884 & 0.1968 & 0.0145 \\
PP\_NR & CD & forehead & exp.\ MA & 1 & 0.3840 & 0.1423 & 0.0068 \\
PP\_NR & CD & forehead & mov.\ avg. & 2 & 0.3418 & 0.1146 & 0.0075 \\
PP\_NR & ED & nose & mov.\ avg. & 1 & 0.3276 & 0.1716 & 0.0064 \\
PP\_NR & ED & nose & median & 2 & 0.3275 & 0.1857 & 0.0019 \\
\bottomrule
\end{tabular}}}
\end{table}

\begin{figure}[ht!]
\vspace{-2mm}
    \centering
    \includegraphics[width=\linewidth]{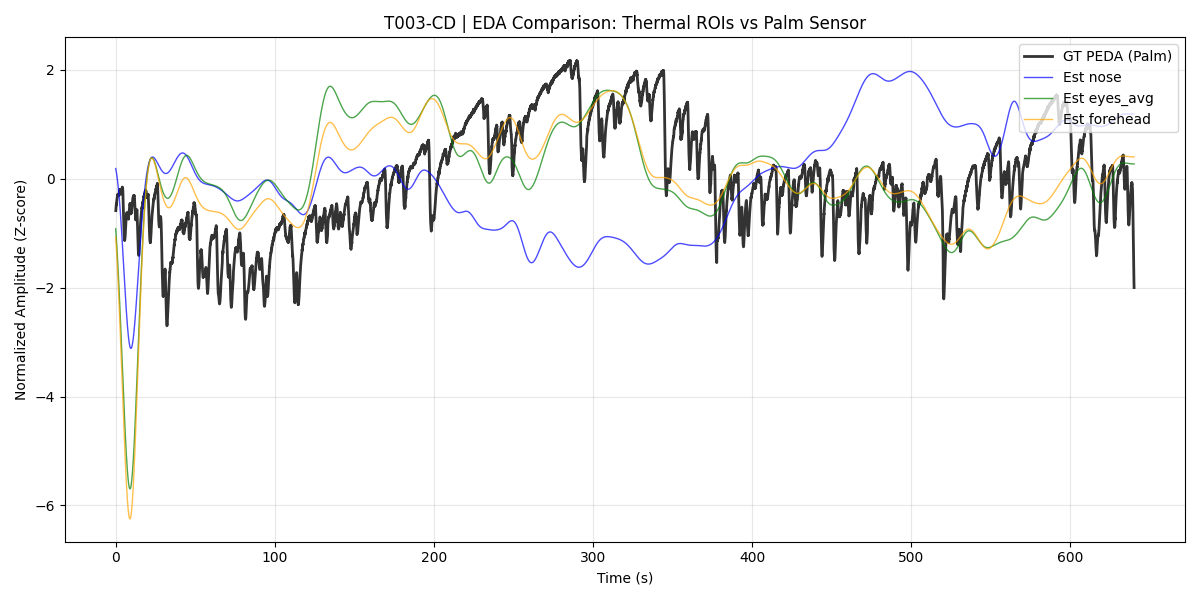}
    \vspace{-7mm}
    \caption{T003-CD: thermal EDA trends from three ROIs vs PEDA ground truth (black).}
    \label{fig:eda_comparison}
    \vspace{-5mm}
\end{figure}

This behavior is also visible in Figure~\ref{fig:eda_comparison} for session T003-CD. The forehead and eye-average trends follow the broad evolution of PEDA more consistently than the nose trace, which deviates between approximately 200 and 400\,s. 


\subsection{Effect of Face Landmarks Smoothing}

\textcolor{black}{The landmark smoothing parameter mainly affects the ROI trace through small spatial jitter. Table~\ref{tab:alpha_sensitivity_small} compares the main fixed EDA baseline with a task-specific case where raw landmarks are less stable. The baseline result is almost unchanged when using raw landmarks ($\alpha=1.0$), showing that the global EDA conclusion is not driven by landmark smoothing. However, in ND for the left periorbital ROI, using raw landmarks reduces the mean $\text{PCC}_{\text{abs}}$ from 0.424 at $\alpha=0.15$ to 0.372. Thus, smoothing is best understood as a robustness step: it has little effect in some aggregate settings, but can prevent degradation for specific task--ROI combinations.}

\vspace{-5mm}
\begin{table}[ht!]
\centering
\def\arraystretch{1.0}
\setlength{\tabcolsep}{2.9em}
\caption{\textcolor{black}{Landmark smoothing sensitivity. Values are mean $\text{PCC}_{\text{abs}}$; $\alpha=1.0$ corresponds to raw landmarks without temporal smoothing. $\Delta_\alpha$ is computed only over smoothed settings.}}
\vspace{-2mm}
\label{tab:alpha_sensitivity_small}
{\color{black}
\begin{tabular}{@{}lcc@{}}
\toprule
\textbf{Smoothing} & \textbf{Global baseline} & \textbf{ND periorbital} \\
\midrule
$\alpha=0.10$ & 0.4029 & 0.4287 \\
$\alpha=0.12$ & 0.4036 & 0.4267 \\
$\alpha=0.15$ & 0.3973 & 0.4240 \\
$\alpha=0.18$ & 0.3814 & 0.4215 \\
$\alpha=0.20$ & 0.3842 & 0.4199 \\
$\alpha=0.30$ & 0.3870 & 0.4113 \\
$\alpha=0.50$ & 0.3932 & 0.3985 \\
$\alpha=1.00$ raw & 0.4054 & 0.3718 \\
\midrule
$\Delta_\alpha$, $\alpha<1.0$ & 0.0222 & 0.0302 \\
\bottomrule
\end{tabular}}

\vspace{1mm}
\begin{minipage}{0.96\linewidth}
\scriptsize
\textcolor{black}{Global baseline: PEDA/global/nose/exp.\ MA. ND periorbital: PEDA/ND/left periorbital/exp.\ MA.}
\end{minipage}
\vspace{-3mm}
\end{table}




\subsection{HR and BR}

The OMIT-based HR pipeline yields MAE $= 13.8 \pm 7.5$\,bpm, bias $= +10.3$\,bpm, and PCC $\approx 0$. These results indicate that cardiac tracking is not reliable in the present setting, where the thermal camera operates at 7.5\,Hz and the subject performs driving tasks with natural head motion. The positive bias suggests that the selected spectral peak may often correspond to motion, respiratory harmonics, or other thermal artifacts rather than the true cardiac component. This does not contradict prior thermal HR studies, but rather highlights the difference between our low-frame-rate driving-simulator setting and previous work using higher frame rates, vessel-focused ROIs, or more controlled poses~\cite{garbey2007contactfree,pereira2018monitoring}.

\begin{table*}[ht!]
\centering
\caption{\textcolor{black}{Comparison with representative thermal physiological studies. Values are reported as in the original publications when available; conditions and protocols differ across studies and are not directly comparable.}}
\vspace{-2mm}
\label{tab:prior_context}
\scriptsize
\def\arraystretch{1.05}
\setlength{\tabcolsep}{0.3em}
{\color{black}
\begin{tabularx}{\textwidth}{@{}p{0.1\textwidth}p{0.15\textwidth}p{0.18\textwidth}p{0.22\textwidth}X@{}}
\toprule
\textbf{Signal} & \textbf{Reference} & \textbf{Setting} & \textbf{Reported result} & \textbf{Relation to this work} \\
\midrule
BR  & Fei \& Pavlidis~\cite{fei2010thermistor}
    & Lab, nostril tracking, thermal waveform
    & High agreement with contact thermistor
    & Foundational virtual-thermistor approach; supports nostril thermal dynamics as a BR cue. \\

BR  & Cho et al.~\cite{cho2017robust}
    & Mobile thermal imaging, dynamic thermal scenes
    & Strong correlation with respiration-belt reference
    & Shows that robust ROI tracking and thermal features improve BR estimation under motion and ambient variation. \\

BR/HR & Pereira et al.~\cite{pereira2018monitoring}
    & Lab, 50\,fps, controlled pose
    & RR RMSE $\approx 0.71$ breaths/min; HR RMSE $\approx 3.5$\,bpm
    & Provides a higher-frame-rate controlled reference; our BR remains usable, while HR degrades in SIM1. \\

\midrule
HR  & Garbey et al.~\cite{garbey2007contactfree}
    & Lab, 30\,fps, vessel-focused thermal ROI
    & Pulse performance 88.52--90.33\%, depending on vessel imprint
    & Indicates that thermal HR requires favorable vascular visibility, tracking, and acquisition conditions. \\

HR  & Pereira et al.~\cite{pereira2018monitoring}
    & Lab, 50\,fps, head-motion proxy
    & HR RMSE $\approx 3.5$\,bpm under controlled acquisition
    & Contrasts with our 7.5\,Hz driving setting, where cardiac peak selection is unreliable. \\

\midrule
EDA & Krzywicki et al.~\cite{krzywicki2014noncontact}
    & High-resolution MWIR, finger and face pore activity
    & Thermal pore activation correlated with skin conductance responses
    & Supports the physiological basis of thermal sudomotor sensing. \\

EDA & Sagaidachnyi et al.~\cite{sagaidachnyi2022separate}
    & Thermal psychophysiology, finger thermograms
    & Separated sweat-gland and hemodynamic temperature components
    & Motivates treating thermal EDA-like trends as a mixture of sudomotor and blood-flow-related effects. \\

EDA/RESP/ PPG & Gioia et al.~\cite{gioia2025thermica}
    & Controlled lab, whole-face ICA
    & Median correlations: EDA $0.9$, respiration $0.8$, PPG envelope $0.73$
    & Provides a controlled-lab reference for multivariate thermal physiology; our driving-simulator mean is lower but peaks at $0.89$. \\

\midrule
\textbf{EDA/BR/HR}
    & This work
    & SIM1 driving simulator, 7.5\,Hz, 6 ROIs
    & BR MAE $3.1\pm1.1$\,bpm; HR MAE $13.8\pm7.5$\,bpm; EDA $\text{PCC}_{\text{abs}}$ $0.40$ mean, $0.89$ peak
    & \textcolor{black}{Common protocol across EDA-like trends, HR, and BR; exposes the performance drop associated with 7.5\,Hz acquisition and driving-task variability.} \\
\bottomrule
\end{tabularx}}
\vspace{-4mm}
\end{table*}

BR estimation is more effective: MAE $= 3.1 \pm 1.1$\,bpm, bias $= +0.2$\,bpm, and PCC $= 0.08 \pm 0.31$. The low PCC should be interpreted with care because window-level breathing rates may have limited dynamic range within each session, while MAE and bias directly quantify the estimation error. The practice drive (PD) yields the best BR performance (MAE $= 2.5$\,bpm, PCC $= 0.32$), whereas cognitive distraction (CD) produces the highest error (MAE $= 3.7$\,bpm), likely due to irregular breathing, head motion, and task-related variability.

\textcolor{black}{Table~\ref{tab:prior_context} provides a protocol-aware comparison with representative thermal physiology studies. The comparison is not a direct leaderboard, since the cited studies differ in camera frame rate, sensor sensitivity, protocol, ROI definition, task constraints, and reference signals. Instead, it makes explicit where the present results fall relative to the literature: BR remains usable but less accurate than controlled high-frame-rate studies, HR degrades substantially in the 7.5\,Hz driving setting, and EDA agreement is lower on average than constrained ICA/high-resolution settings but can reach comparable values in individual sessions.}

\subsection{Task, Subject, and Demographic Effects}

\begin{figure*}[ht!]
\vspace{-2mm}
    \centering
    \includegraphics[width=\linewidth]{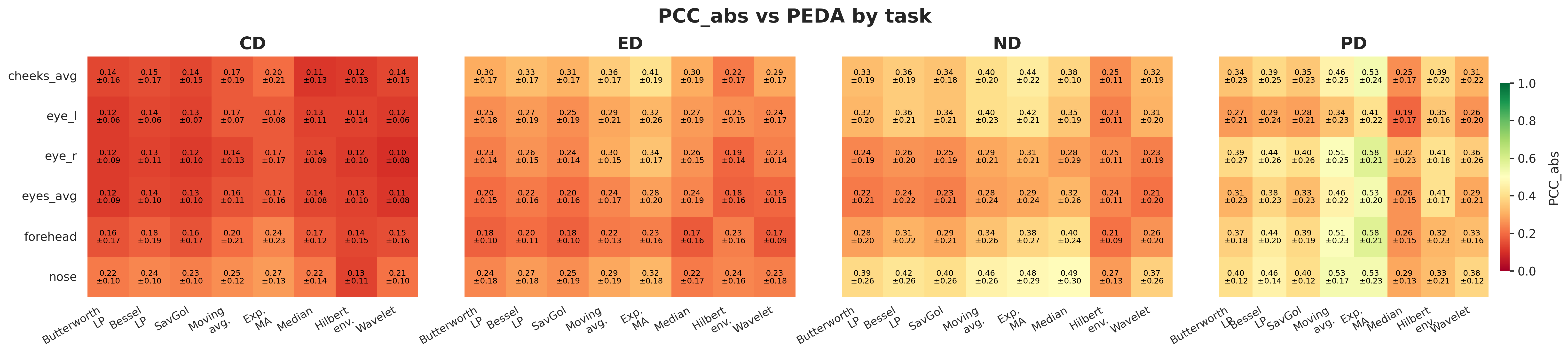}
    \vspace{-8mm}
    \caption{$\text{PCC}_{\text{abs}}$ against PEDA by driving condition ($n = 7$--$8$ per task).}
    \label{fig:heatmap_by_task}
    \vspace{-2mm}
\end{figure*}

Figure~\ref{fig:heatmap_by_task} shows estimated EDA against PEDA by condition. Normal driving (ND) yields the most stable PP\_NR agreement (cheeks/exp.\ MA: $0.42 \pm 0.09$). Cognitive distraction shifts the best ROI to the forehead ($0.38 \pm 0.15$), and PD produces the highest PEDA correlations ($0.58 \pm 0.21$).

Table~\ref{tab:subject_results} reports per-subject EDA. The best $\text{PCC}_{\text{abs}}$ ranges from 0.26 to 0.44 (PP\_NR) and 0.35 to 0.67 (PEDA), with the preferred ROI varying across subjects. Table~\ref{tab:demographics} shows demographic effects. For EDA, both sexes produce similar PEDA correlations (0.31), while older subjects show slightly higher values (0.32 vs 0.30). For HR, younger subjects achieve lower MAE (8.9 vs 18.3\,bpm), suggesting that age-related skin properties may affect cardiac signal quality. BR performance is consistent across groups (MAE $\approx 3.0$--$3.2$\,bpm). These observations are preliminary given the small sample.

\begin{table}[ht!]
\centering
\def\arraystretch{1.1}
\setlength{\tabcolsep}{0.7em}
\caption{Best EDA ROI-method per subject (mean across sessions - n).}
\vspace{-3mm}
\label{tab:subject_results}
\resizebox{\linewidth}{!}{%
\begin{tabular}{@{}lcccll@{}}
\toprule
\textbf{Subj.} & \textbf{S/A} & $n$ & \textbf{Best (PP\_NR)} & \textbf{Best (PEDA)} \\
\midrule
T002 & F/Y & 4 & cheeks/eMA 0.44$\pm$0.19 & nose/med 0.36$\pm$0.22 \\
T003 & M/Y & 4 & nose/mAvg 0.35$\pm$0.23 & cheeks/eMA 0.53$\pm$0.16 \\
T005 & M/Y & 3 & foreh./med 0.39$\pm$0.09 & cheeks/eMA 0.35$\pm$0.05 \\
T014 & F/Y & 4 & cheeks/eMA 0.43$\pm$0.19 & cheeks/eMA 0.39$\pm$0.34 \\
T029 & F/O & 4 & nose/mAvg 0.40$\pm$0.15 & cheeks/eMA \textbf{0.67$\pm$0.21} \\
T031 & F/O & 4 & cheeks/eMA 0.26$\pm$0.18 & eye\_l/eMA 0.45$\pm$0.23 \\
T034 & M/O & 4 & nose/mAvg \textbf{0.44$\pm$0.28} & eye\_l/eMA 0.42$\pm$0.25 \\
T036 & M/O & 4 & nose/eMA 0.38$\pm$0.14 & nose/mAvg 0.43$\pm$0.21 \\
\midrule
\multicolumn{3}{@{}l}{\textit{Mean}} & 0.39$\pm$0.06 & 0.45$\pm$0.10 \\
\bottomrule
\end{tabular}}
\vspace{-1mm}
\end{table}

\begin{table}[ht!]
\centering
\def\arraystretch{1.05}
\caption{Demographic effects: EDA ($\text{PCC}_{\text{abs}}$, Butterworth LP, nose), HR and BR (MAE, bpm).}
\vspace{-3mm}
\label{tab:demographics}
\scriptsize
\resizebox{\linewidth}{!}{%
\begin{tabular}{@{}lccccc@{}}
\toprule
\textbf{Group} & $n$ & \textbf{EDA vs PP} & \textbf{EDA vs PEDA} & \textbf{HR MAE} & \textbf{BR MAE} \\
\midrule
Female   & 16 & 0.22$\pm$0.14 & 0.31$\pm$0.18 & 15.1$\pm$9.7 & 3.0$\pm$2.0 \\
Male     & 15 & 0.33$\pm$0.19 & 0.31$\pm$0.20 & 12.3$\pm$8.9 & 3.2$\pm$2.1 \\
\midrule
Young    & 15 & 0.25$\pm$0.16 & 0.30$\pm$0.23 & ~~8.9$\pm$8.7 & 3.0$\pm$1.9 \\
Older    & 16 & 0.28$\pm$0.19 & 0.32$\pm$0.14 & 18.3$\pm$9.9 & 3.2$\pm$2.2 \\
\bottomrule
\end{tabular}}
\vspace{-3mm}
\end{table}

\subsection{Latency, Polarity, and Signal Examples}

\begin{figure}[ht!]
    \centering
    \includegraphics[width=\linewidth]{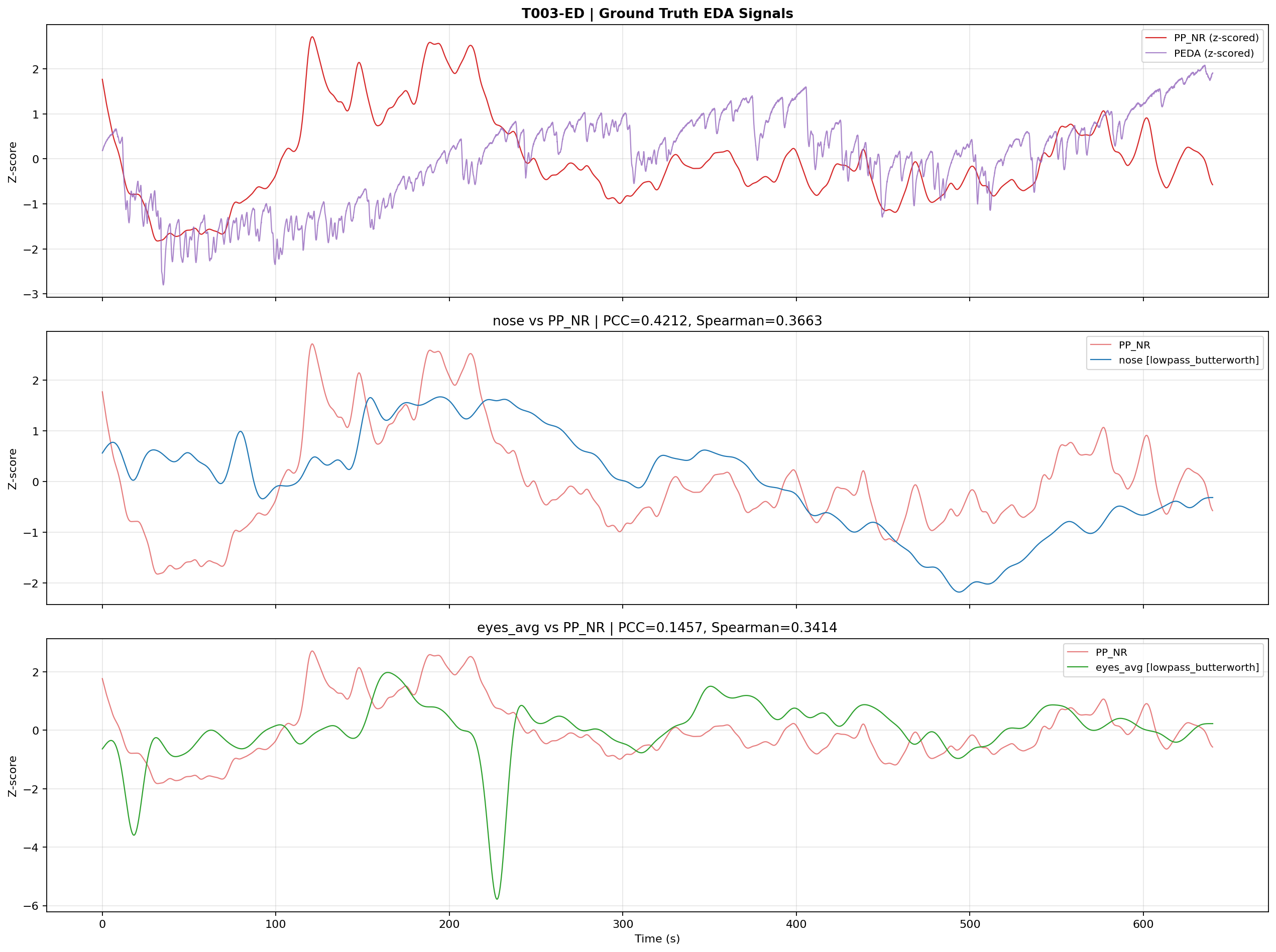}
    \vspace{-5mm}
    \caption{T003-ED: PP\_NR and PEDA GT (top), nose ROI vs PP\_NR (middle, $r=0.42$), eyes ROI (bottom, $r=0.15$).}
    \label{fig:timeseries_example}
\end{figure}

The cross-correlation analysis yields a median lag of 0.0\,s for both references, with mean absolute lag $\approx 19$--$20$\,s ($\sigma \approx 32$\,s). Of 31 sessions, 55\% (PP\_NR) and 77\% (PEDA) peak within $|\tau| < 30$\,s. For well-tracked signals, the latency is typically below 15\,s. \textcolor{black}{Because most peaks fall within 30\,s and the median lag is zero, real-time deployment can rely on short causal delay compensation rather than the offline $\pm 120$\,s search.} The thermal-EDA polarity alternates across sessions: 58\% positive against PP\_NR, 52\% against PEDA. This near-chance split motivates $\text{PCC}_{\text{abs}}$ and indicates that practical systems must handle polarity inversions. The alternation reflects competing vasoconstriction (cooling) and eccrine sweating (warming) mechanisms. \textcolor{black}{Table~\ref{tab:polarity_signed_vs_abs} contrasts the signed and absolute Pearson correlations across the 31 sessions: averaging the signed value at the population level collapses real agreement toward zero, whereas $\text{PCC}_{\text{abs}}$ retains it.} Figure~\ref{fig:timeseries_example} shows session T003-ED. The nose trend (Butterworth LP) tracks the broad PP\_NR shape (PCC $= 0.42$), while the eyes-average signal (PCC $= 0.15$) shows less agreement and a tracking artifact near $t = 250$\,s.

\begin{table}[ht!]
\centering
\def\arraystretch{1.05}
\setlength{\tabcolsep}{0.7em}
\caption{\textcolor{black}{Signed vs.\ absolute Pearson correlation across 31 sessions for the best fixed configuration. The near-50/50 polarity split makes the signed mean uninformative at the population level, while $\text{PCC}_{\text{abs}}$ recovers the underlying agreement.}}
\vspace{-3mm}
\label{tab:polarity_signed_vs_abs}
\scriptsize
\resizebox{\linewidth}{!}{%
\textcolor{black}{%
\begin{tabular}{@{}lcccc@{}}
\toprule
\textbf{Reference} & \textbf{Mean signed PCC} & \textbf{Mean $\text{PCC}_{\text{abs}}$} & \textbf{\% positive} & \textbf{\% negative} \\
\midrule
PP\_NR & \textcolor{black}{$\approx 0$} & 0.31 & 58 & 42 \\
PEDA   & \textcolor{black}{$\approx 0$} & 0.40 & 52 & 48 \\
\bottomrule
\end{tabular}}}
\end{table}

%
%
%
%

\section{Discussion and Conclusions}\label{sec:conclusions}

\textcolor{black}{This study characterizes the contactless extraction of sudomotor (sweat-gland-related) trends, cardiac pulse, and respiratory rate from facial thermal video, using 288 EDA configurations across 31 SIM1 sessions together with multi-ROI HR and BR pipelines. The results indicate which thermal cues can be recovered under low-frame-rate, motion-affected, and task-dependent conditions.}

\textcolor{black}{For EDA-like trends, the nose and cheeks with time-domain smoothers give the strongest agreement (mean $\text{PCC}_{\text{abs}}=0.40$ against palm EDA, session peaks up to $0.89$). The small differences between nose, cheeks, and forehead and the task- and session-dependent ROI preferences indicate that facial regions capture partly different autonomic manifestations, supporting ROI-confidence estimation or multi-ROI fusion rather than a single fixed region for adaptive HMI. The near-balanced polarity split further shows that these trends should be handled with polarity-aware or polarity-invariant representations.}

\textcolor{black}{The cardiorespiratory results differ in reliability: BR reaches $3.1$\,bpm MAE with near-zero bias and remains usable at 7.5\,Hz, whereas HR yields $13.8$\,bpm MAE and no meaningful temporal correlation, limited by weak thermal modulation, motion, and the low frame rate. This suggests a hierarchy of thermal cues for adaptive HMI: respiration is the most stable low-rate input, EDA-like trends can add arousal or workload context once ROI quality and polarity are handled, and HR requires higher frame rates, vessel-specific ROIs, or complementary sensing.}

\textcolor{black}{Demographic analyses add further variability: HR errors are lower for younger than older subjects, while EDA and BR are more stable across groups, possibly reflecting age-dependent skin and vascular properties, though larger cohorts are needed for confirmation. Together with condition-dependent ROI preferences, this indicates that fixed pipelines may be insufficient for deployment, motivating subject-specific calibration and adaptive fusion.}

\textcolor{black}{The main limitations are the small balanced subset used for subgroup analysis, the low thermal frame rate for HR estimation, the focus on trend-level rather than event-level EDA evaluation, and the absence of explicit testing under occlusions that block or alter infrared emission. Eyeglasses can mask the periorbital region, while dense facial hair, masks, and industrial PPE can decouple measured thermal traces from the underlying skin temperature. Since SIM1 was not designed to evaluate these factors, robustness to occlusions, airflow, changing ambient temperature, and real industrial protective equipment remains an important direction for future work.}

\textcolor{black}{Future studies should evaluate higher-frame-rate thermal cameras, adaptive multi-ROI fusion driven by signal-quality indicators, subject-specific calibration for sudomotor extraction, and learning-based fusion methods benchmarked against the present interpretable protocol. Validation should also move toward vehicle cabins, machinery operator stations, and human-robot collaboration settings, where edge-based processing may support privacy-preserving operator-state monitoring.} In general, the results provide baseline performance limits and design guidance for thermal contactless biosignal extraction as an auxiliary sensing layer in adaptive industrial HMI systems.

\section*{Acknowledgment}

\textcolor{black}{This research was supported by the Research Council of Finland, formerly the Academy of Finland, through the 6G Flagship Programme under Grant 369116, and Profi7 Hybrid Intelligence programme (352788), and by the Business Finland 6G-WISECOM project (Grant 3630/31/2024). The authors wish to acknowledge the use of AI-assisted tools for language editing, with a focus on grammar checking and improving readability, and coding support. In addition, the Elicit platform was used during the literature search as an AI-assisted discovery aid for candidate records. All inclusion decisions, exclusions, data extraction, synthesis, interpretation, and final writing decisions were performed by the authors.}


\noindent

\bibliographystyle{IEEEtran}
\bibliography{references}

\end{document}